\documentclass{article}

\usepackage[preprint]{neurips_2025}

\usepackage[utf8]{inputenc} 
\usepackage[T1]{fontenc}    
\usepackage{hyperref}       
\usepackage{url}            
\usepackage{booktabs}       
\usepackage{amsfonts}       
\usepackage{nicefrac}       
\usepackage{microtype}      
\usepackage{xcolor}         

\usepackage{algorithm}
\usepackage{algorithmic}
\usepackage{amsmath}	
\usepackage{amssymb}	
\usepackage{times}
\usepackage{epsfig}
\usepackage{float}
\usepackage{placeins}
\usepackage{color, colortbl}
\usepackage{stfloats}
\usepackage{enumitem}
\usepackage{tabularx}
\usepackage{xstring}
\usepackage{multirow}
\usepackage{xspace}
\usepackage{url}
\usepackage[hang,flushmargin]{footmisc}
\usepackage{svg}
\usepackage{graphicx}
\usepackage[normalem]{ulem}
\usepackage{cleveref}

\useunder{\uline}{\ul}{}

\NewDocumentCommand{\tcite}{m}{~\cite{#1}\xspace}

\newcommand{\simbolo}[2]{%
    \ifcsname mycounter#1\endcsname
    \else
        \newcounter{mycounter#1}%
    \fi
    \expandafter\newcommand\csname#1\endcsname[1][]{%
            \addtocounter{mycounter#1}{1}%
            \ifnum\value{mycounter#1}>1%
                {#1}{##1}\ifblank{##1}{}{\noexpand}\xspace%
            \else%
                {#2{##1} (#1{##1})}\ifblank{##1}{}{\noexpand}\xspace%
            \fi%
    }%
}

\simbolo{MFIF}{Multi-Focus Image Fusion}
\simbolo{DCT}{Discrete Cosine Transform}
\simbolo{DL}{Deep Learning}
\simbolo{ML}{Machine Learning}
\simbolo{CV}{Computer Vision}
\simbolo{AI}{Artificial Intelligence}
\simbolo{CNN}{Convolutional Neural Network}
\simbolo{DNN}{Deep Neural Network}
\simbolo{FC}{Fully Connected}
\simbolo{MSE}{Mean Squared Error}
\simbolo{GT}{Ground Truth}
\simbolo{VAE}{Variational Autoencoder}
\simbolo{HDR}{High Dynamic Range}
\simbolo{DSE}{Defocus Spread Effect}
\simbolo{DOF}{Depth-of-Field}

\title{Addressing the Depth-of-Field Constraint: A New Paradigm for High Resolution Multi-Focus Image Fusion}

\author{%
    Luca Piano\textsuperscript{1,2} \qquad 
    Peng Huanwen\textsuperscript{1} \qquad
    Radu Ciprian Bilcu\textsuperscript{1} \\
    Huawei Technologies Oy (Finland) Company Ltd\textsuperscript{1}, 
      Politecnico di Torino\textsuperscript{2}\\
    \texttt{\{luca.piano\}@polito.it}\\
    \texttt{\{penghuanwen\}@huawei.com}
}

\begin{document}

\maketitle

\begin{abstract}
Multi-focus image fusion (MFIF) addresses the depth-of-field (DOF) limitations of optical lenses, where only objects within a specific range appear sharp. Although traditional and deep learning methods have advanced the field, challenges persist, including limited training data, domain gaps from synthetic datasets, and difficulties with regions lacking information.
We propose VAEEDOF, a novel MFIF method that uses a distilled variational autoencoder for high-fidelity, efficient image reconstruction. Our fusion module processes up to seven images simultaneously, enabling robust fusion across diverse focus points. To address data scarcity, we introduce MattingMFIF, a new syntetic 4K dataset, simulating realistic DOF effects from real photographs.
Our method achieves state-of-the-art results, generating seamless artifact-free fused images and bridging the gap between synthetic and real-world scenarios, offering a significant step forward in addressing complex MFIF challenges. The code, and weights are available here: \codelink{}
\end{abstract}
\section{Introduction}
\label{sec:intro}



\MFIF{} is a significant challenge in image processing. Optical lenses are inherently limited by the \DOF constraint, where only objects within a certain distance appear sharp, while objects outside this range become blurred. As a result, it is difficult to capture a scene in which all objects at varying distances are simultaneously in focus in a single shot\tcite{aslantas2010fusion}. To address this issue, various algorithms have been proposed to fuse multiple images with different focus points, producing an all-in-focus image. These fused images can enhance visualization, enable more effective feature extraction, and improve object recognition in computer vision tasks.

While traditional methods have provided valuable solutions, they often have inherent limitations that hinder their effectiveness in more complex scenarios. In recent years, methods based on \DL have gained significant attention as they offer greater flexibility and potential for improved fusion results. Despite their promising performance, they still present several unresolved challenges. One key issue is the lack of sufficient and diverse training data. Current publicly available datasets are limited in size and resolution, and synthetic data generation techniques often introduce domain gaps with respect to real images. Moreover, these datasets tend to lack challenging cases, which are crucial for developing robust models.

Furthermore, the nature of the models themselves contributes to the difficulty of the problem. Decision mask-based approaches, which focus on selecting the sharpest regions for fusion, face challenges in dealing with areas where information is missing, for instance, when foreground elements block the view of background objects. However, current end-to-end methods struggle to produce high-quality fused images due to the limited number of training examples, making them prone to generating artifacts and failing to generalize well to real-world scenarios.

\begin{figure}[tbh]
    \centering
    \includegraphics[width=0.8\columnwidth]{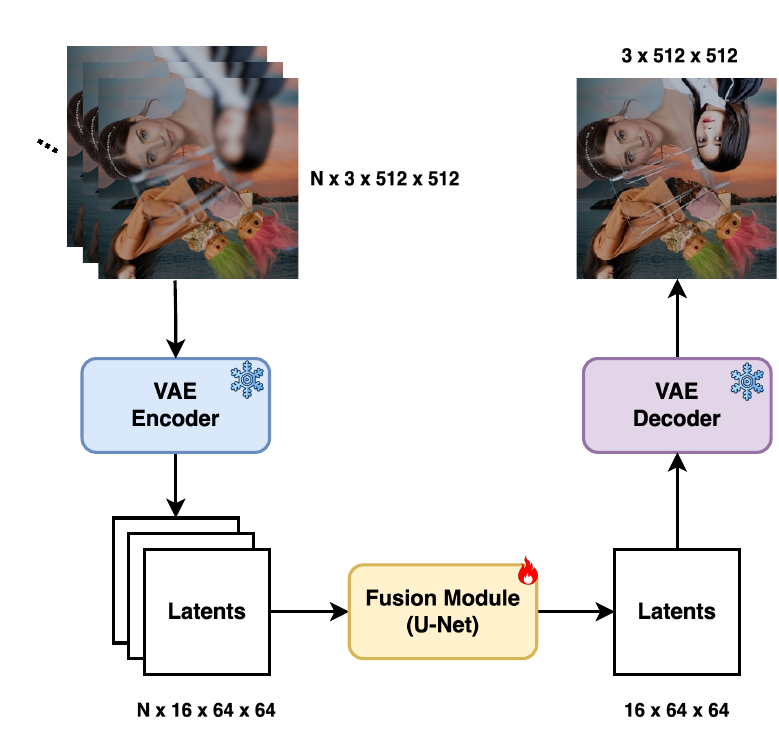}
    \caption{Overview of the architecture. VAE is pretrained and frozen during the training.}
    \label{fig:architecture}
\end{figure}

In this paper, we propose VAEEDOF to overcome these challenges. Our method leverages a \VAE distilled from Stable Diffusion 3 to balance high-fidelity image reconstruction with efficient speed. We introduce a Fusion U-Net module capable of fusing up to seven different images simultaneously, allowing for more comprehensive fusion across multiple focus points. Furthermore, we present a new 4k \MFIF training and validation dataset created using real photos, simulating \DOF effects with Blender. This dataset provides a realistic large-scale foundation for training models that addresses the scarcity of high-quality data for \MFIF. Our approach aims to improve the quality of fused images, bridge the domain gap, and handle complex scenarios effectively.

\section{Related Work}
\label{sec:related}

This section initially delves into conventional techniques for \MFIF. Following that, the discussion shifts to approaches rooted in \DL methods and, subsequently, an evaluation of the existing dataset is conducted.

\subsection{Multi-Focus Image Fusion}

\MFIF was initially introduced in the late 1970s, leading to the development of numerous methods aimed at addressing this particular task, which can be broadly categorized into traditional methods and \DL-based approaches. The traditional methods for \MFIF are further divided into two main categories: transform- and spatial domain-based techniques.
Classical transform domain fusion methods rely on multiscale transforms, including the Laplacian Pyramid, and the Discrete Wavelet Transform, among others. These methods first convert images into a different domain, perform fusion on the transformed coefficients, and then reconstruct the fused image using the inverse transform \tcite{ma2019multi,ma2017multi,qiu2019guided,liu2015multi,shreyamsha2015image,tian2011multi,bai2015quadtree,liu2020multi}. In contrast, spatial domain-based methods focus on processing image pixels to extract and fuse relevant information. Pixel-based methods involve generating decision maps based on sharpness or focus criteria, which are refined using various filtering and comparison techniques\tcite{ma2019multi,ma2017multi,qiu2019guided,liu2015multi,shreyamsha2015image,tian2011multi,bai2015quadtree}. Block-based methods decompose source images into blocks and use specific measures to detect focussed regions from which the fused image is constructed\tcite{bai2015quadtree}. Region-based methods first identify the boundaries between the focussed and defocussed regions, segment the image accordingly, and calculate focus measures to generate a decision map for fusion\tcite{zhang2017boundary}.



Although traditional methods provide a solid foundation for \MFIF, \DL-based techniques are gaining increasing attention due to their ability to learn complex features and fusion strategies from data. These techniques adeptly capture intricate data patterns, allowing effective integration of multi-focus information. These techniques can be broadly classified into decision map-based and generative methods\tcite{zhang2021deep}. The former use decision map generation and potential post-processing for refinement. The fused image is derived by blending the first image ($I_0$) and the second image ($I_1$) according to the decision mask ($M$), using the formula: $I_0  M + I_1 (1-M)$

Liu et al.\tcite{liu2017multi} introduced one of the first \DL-based methods, which uses a \CNN to directly learn the mapping from the source images to the focus map. Then, a large number of scholar have joint the research.  SESF-Fuse\tcite{ma2021sesf} trains an encoder–decoder network in an unsupervised manner to extract deep features from input images. The spatial frequency is then used to measure the activity levels of these features, followed by consistency verification to refine the decision map and generate the fused result. Wang et al.\tcite{wang2021mfif} introduce MFIF-GAN to mitigate \DSE by generating focus maps with accurately sized foreground regions. It incorporates reconstruction and gradient regularization terms in the loss function to enhance boundary details and improve fusion quality. GACN\tcite{ma2022end} uses a cascade network trained end-to-end, eliminating the reliance on empirical postprocessing. To further enhance the output, it employs a gradient-aware loss function to preserve fine gradient details in the fused image. MSI DTrans\tcite{zhai2024msi} is a fusion method that combine multilayer semantic interaction with dynamic transformers. Integrates Haar wavelets into the network feature extraction process to combine the strengths of traditional and \DL methods, enabling more comprehensive feature representation. 
Although these models have yielded promising results, they have a significant drawback: they cannot handle regions with missing information in the input images.

In contrast, generative methods bypass the decision map, processing source images directly to output the fused image. IFCNN\cite{zhang2020ifcnn} draws inspiration from transform-domain fusion algorithms, using convolutional layers to extract salient features from input images, which are then fused using different rules. The fused features are reconstructed through additional convolutional layers in an end-to-end fashion, eliminating the need for post-processing.
Ma et al.\tcite{ma2022swinfusion} proposed SwinFusion, a general image fusion framework leveraging cross-domain long-range learning with the Swin Transformer. Complementary information is integrated through attention-guided fusion units based on self-attention and cross-attention mechanisms, while utilizing shifted windows for flexible image sizes. FusionDiff\tcite{li2024fusiondiff} was the first diffusion-based model for \MFIF, achieving good results with a small training set. Aiming to combine the strengths of generative and  decision map-based methods, Zhang et al.\tcite{zhang2024exploit} propose a dual-branch network providing a robust solution for \MFIF.

The generative nature of this class of models has the potential to address the limitations of masking-based methods. However, they also introduce the risk of generating artifacts in regions where masking methods typically perform well. This issue is further exacerbated by the scarcity of large, high-quality datasets required for training models that aim to fully reconstruct images.

\subsection{Datasets}

Supervised deep learning algorithms typically require large, high-quality labeled datasets for effective training. However, existing \MFIF datasets are often limited in size and primarily intended for testing rather than training. Furthermore, creating \GT for complex or challenging cases, or simulating these cases, is particularly difficult.  Two major real datasets are created using light field data: the Lytro dataset \cite{nejati2015multi} and the Real-MFF dataset \cite{zhang2020real}. Building such datasets requires considerable human effort and they are limited in size, which makes them primarily suitable for testing purposes. Furthermore, these datasets are often limited in resolution. To address this issue, Xiao et al. captured a high-resolution multi-focus image dataset, termed HighMF\tcite{xiao2022dmdn}. However, this dataset lacks \GT data, restricting its applicability exclusively to visual testing scenarios.


To create datasets for training, many prior works have relied on synthetic data. Researchers\tcite{liu2017multi,luo2024review} often use point spread function-based methods to generate defocused regions for the synthesis of datasets. However, this approach relies on accurately identifying defocused areas, creating a gap between synthetic datasets and real-world scenarios, which remain challenging to replicate\tcite{luo2024review}.

Chen et al.\tcite{chen2024defocus} propose using 3D generated images with Blender Cycles to create more realistic datasets. However, a domain gap between 3D-generated and real images may undermine performance when applied to real-world tasks\tcite{piano2023bent,ljungqvist2023object}.

\section{Method}
\label{sec:method}

We propose a new architecture to address two key challenges of existing MFIF methods: low-quality results in regions where input images are lacking information and the artifacts in regions where input images have different distributions. 

\begin{figure}[tbh]
    \centering
    \includegraphics[width=\textwidth]{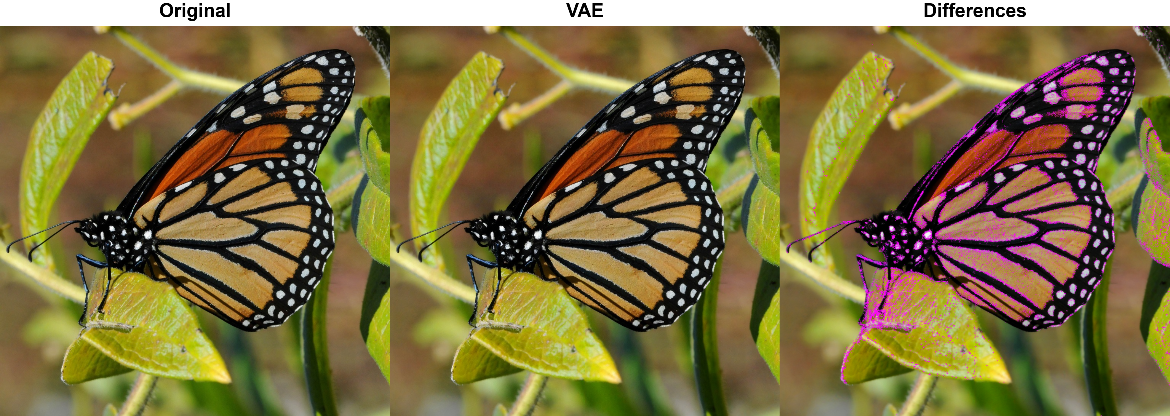}
    \caption{The figure shows the differences between the original and the reconstructed image, with discrepancies highlighted in purple.}
    \label{fig:vae_err}
\end{figure}

The proposed architecture (\Cref{fig:architecture}) consists of two main components. The first component is a frozen pre-trained \VAE, which operates in a compressed latent feature space. We analyzed different \VAE models and ultimately selected a 16-channel \VAE distilled from StableDiffusion3\footnote{\url{https://github.com/madebyollin/taesd}}. The \VAE allows the architecture to bypass the need to relearn pixel generation, ensuring more stable and accurate reconstructions (\Cref{fig:vae_err}).

The second component of the architecture is the Fusion Module (FM), which is responsible for the fusion process. The module is a U-Net network composed of three downsampling and upsampling blocks. Each block is fundamentally a residual block, and the middle blocks are further enhanced with an attention layer.

The model was trained for 180,000 steps on four NVIDIA V100 GPUs using 512×512 image patches and a global batch size of 32. For inference, we use the exponential moving average (EMA) of the model weights, which enhances stability and improves generation quality.

Seven distinct images are encoded separatly by the \VAE encoder, then concatenated and fed into the FM, which produces a fused latent representation. This fused representation is then decoded by the frozen \VAE decoder to generate the final fused image. In cases where fewer than seven images are available, existing images can be duplicated to ensure that the FM receives the required input dimensions.

\begin{figure}[htb]
    \centering
    \includegraphics[width=0.45\columnwidth]{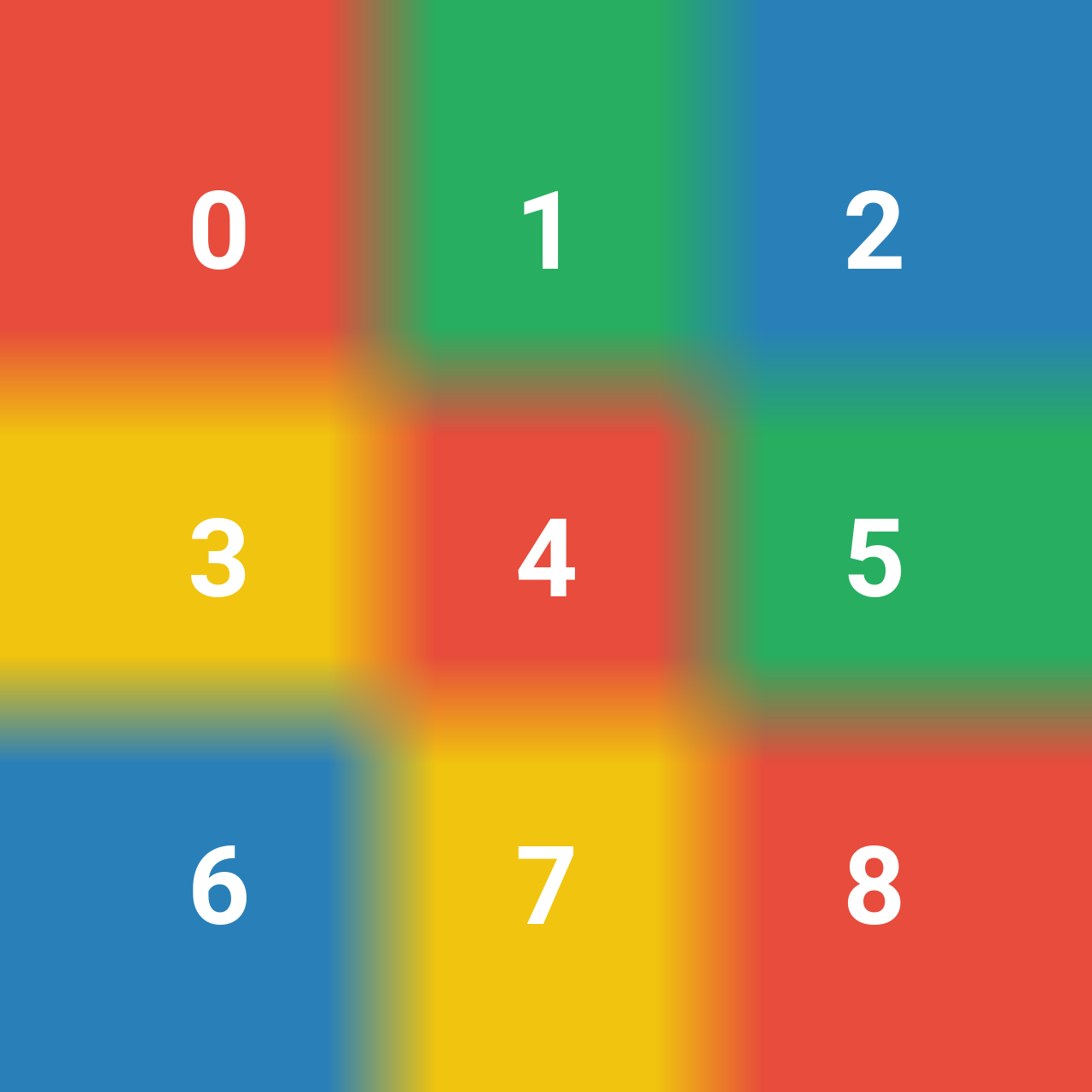} 
    \includegraphics[width=0.45\columnwidth]{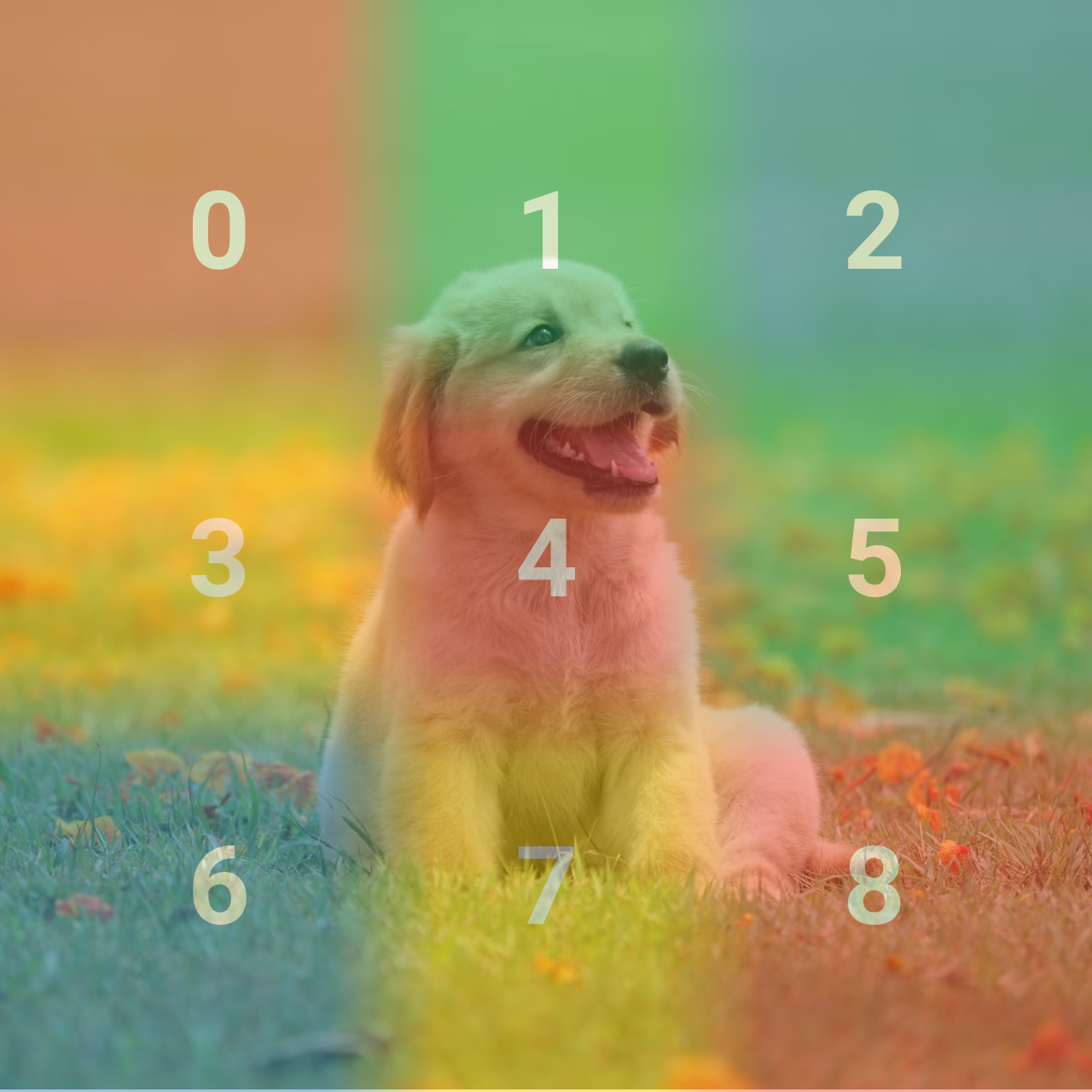} 
    \caption{Visualization of the patch combination process.}
    \label{fig:alpha_mask}
\end{figure}

To enhance the quality and eliminate artifacts at the boundaries of the image patches, an overlapping strategy is used. Rather than processing disjoint patches, in our approach, each patch overlaps with its adjacent patch by $\frac{1}{4}$ of the patch size. Across these overlapping regions, a gradient alpha mask is applied to ensure seamless transitions and minimize conjunction artifacts (\Cref{fig:alpha_mask}). This strategy improves visual coherence and addresses challenges that arise in complex scenes, where maintaining consistency across patches is essential for high-quality fusion results.

To enable high quality \MFIF, we propose a novel dataset generated using Blender (MattingMFIF). Unlike previous works that aim for highly photorealistic 3D object renders, we simplify the problem by focusing on a controlled scene setup. Specifically, we used cropped subjects from the Distinctions-646\tcite{qiao2020attention} and PhotoMatte85\tcite{lin2021real} datasets, along with background images from BK-20K\tcite{li2022bridging}. To generate various \MFIF examples, a Blender script (\Cref{alg:the_alg}) is utilized to accurately position a camera and six distinct planes at various distances from the camera's viewpoint. For the first five planes, we applied one of the subjects as a texture, randomly adjusting both the position and size of the plane. The last plane is positioned farther from the others and resized to fill the entire field of view. Then we generate a render for each subject, adjusting the focus to ensure that only one is in focus in each render. The casting of shadows was disabled during the rendering process to prevent the introduction of unrealistic shadows. 

\begin{algorithm}[h]
\caption{Dataset generator algorithm}
\begin{algorithmic}[1]
\label{alg:the_alg}
\STATE $n \gets 5$
\STATE Initialize Blender scene settings.
\STATE Add a new camera to the scene.
\STATE Set camera properties:
\STATE \quad Enable Depth Of Field
\STATE \quad $\text{f-stop} \gets \text{random}(0.1, 1.5)$
\STATE \quad $\text{focus-length} \gets \text{random}(90, 120)$ mm.
\STATE Set render settings:
\STATE \quad Set render resolution $(4096,4096)$
\STATE \quad Disable shadows
\STATE \quad Set random HDR image.
\FOR{$i \gets 1$ to $n$}
    \STATE Apply random position and scale.
    \STATE Distance from camera in $[1,9.5]$m.
    \STATE Apply subject image as texture.
    \STATE Apply random rotation.
\ENDFOR
\STATE Add a plane at 10m.
\STATE Scale to fill the FOV.
\STATE Set background image as texture.
\STATE Render each scene with different focus distances:
\FOR{$i \gets 1$ to $n + 1$}
    \STATE Set focus distance to the distance of the $i$-th plane.
    \STATE Render image.
\ENDFOR
\STATE Disable depth of field.
\STATE Render a GT.
\end{algorithmic}
\end{algorithm}

Lastly, \GT is created by generating a render with the DOF disabled, ensuring a sharp focus across the entire scene. This approach yields images that more closely resemble real-world scenarios, as the objects are captured from cameras and the renders simulate the natural blurring seen in reality. Not only does this method provide a realistic \DSE, it also produces a perfect \GT, which would otherwise be difficult to achieve. To further enhance diversity in the dataset, 10 different \HDR images are randomly selected to modify the lighting conditions during the rendering process. We generate 10,000 samples for training, ensuring a diverse set of fusion challenges. For the test set, we used the 200 images from the AM-2k\tcite{li2022bridging} evaluation dataset without repetitions. As background images, we randomly select from the LIU4K-v2\tcite{Liu4K} dataset, focusing exclusively on images classified as buildings and mountains.

\begin{figure}[!h]
    \centering
    \includegraphics[width=\columnwidth]{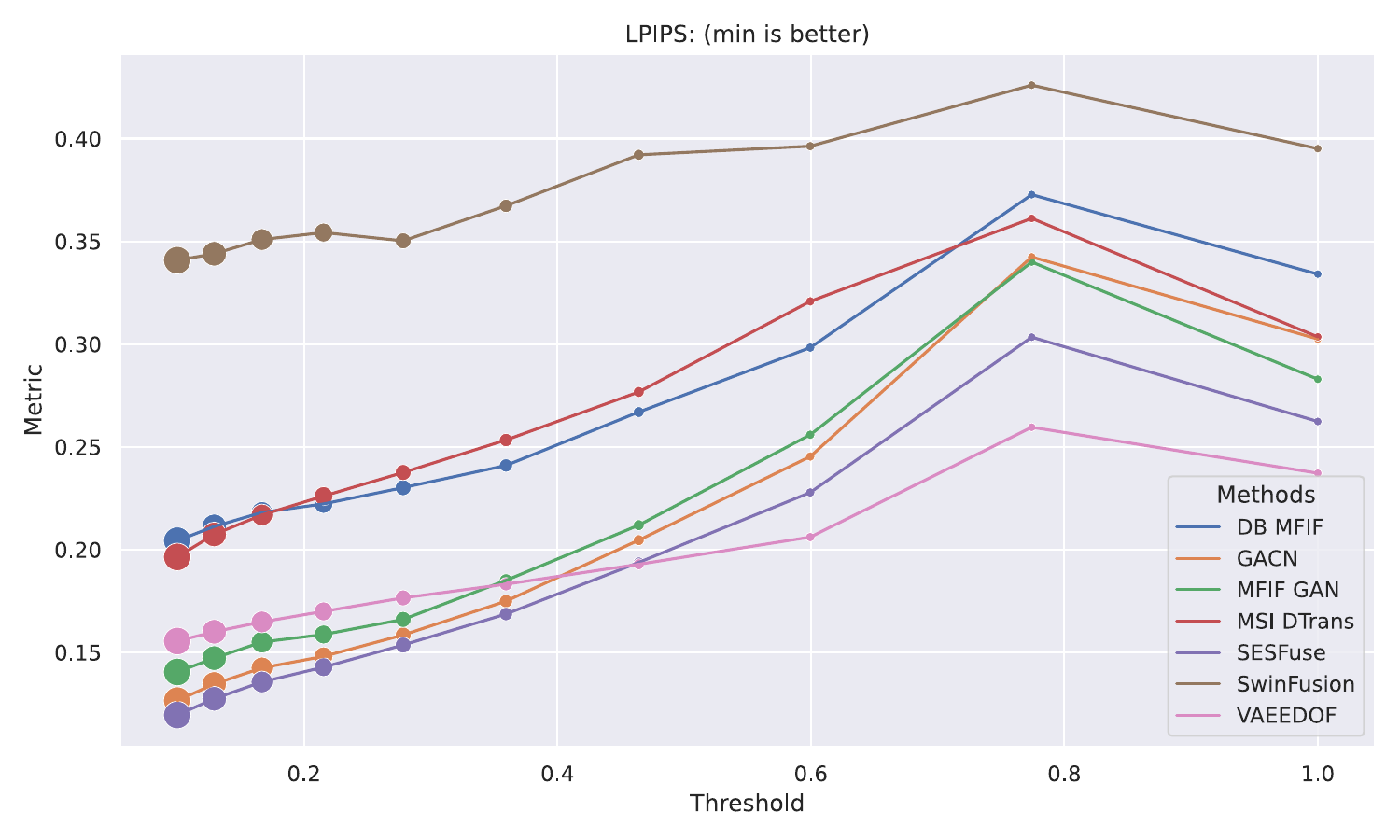}
    \caption{The plot shows the LPSIS metric for different image patches categorized by their difficulty level. The size of each point represents the number of patches within a specific difficulty threshold, with higher thresholds corresponding to more challenging regions.}
    \label{fig:scores}
\end{figure}

\section{Results}
\label{sec:results}

\begin{table*}[t]
\caption{Comparison of our method with the state-of-the-art methods on our MattingMFIF. $\uparrow$ indicates higher is better, while  $\downarrow$ i indicates lower is better.}
\label{tab:results}
\centering
\resizebox{\textwidth}{!}{%
\begin{tabular}{cccccccccccc}
\toprule
 &
  \multicolumn{11}{c}{{Metrics}} \\ \cmidrule{2-12}
{Methods} &
  \multicolumn{1}{c}{{DISTS $\downarrow$}} &
  \multicolumn{1}{c}{{PieAPP $\downarrow$}} &
  \multicolumn{1}{c}{{HaarPSI $\uparrow$}} &
  \multicolumn{1}{c}{{IW-SSIM $\uparrow$}} &
  \multicolumn{1}{c}{{LPIPS $\downarrow$}} &
  \multicolumn{1}{c}{{MDSI $\downarrow$}} &
  \multicolumn{1}{c}{{MSE $\downarrow$}} &
  \multicolumn{1}{c}{{MS-SSIM $\uparrow$}} &
  \multicolumn{1}{c}{{MS-GMSD $\downarrow$}} &
  \multicolumn{1}{c}{{PSNR$\uparrow$}} &
  \textbf{SSIM $\uparrow$} \\ \midrule
{DB MFIF} &
  \multicolumn{1}{c}{0.071} &
  \multicolumn{1}{c}{1.289} &
  \multicolumn{1}{c}{0.783} &
  \multicolumn{1}{c}{0.924} &
  \multicolumn{1}{c}{0.094} &
  \multicolumn{1}{c}{0.300} &
  \multicolumn{1}{c}{{\ul 0.002}} &
  \multicolumn{1}{c}{0.949} &
  \multicolumn{1}{c}{0.083} &
  \multicolumn{1}{c}{28.170} &
  0.945 \\
{GACN} &
  \multicolumn{1}{c}{0.042} &
  \multicolumn{1}{c}{{\ul 0.593}} &
  \multicolumn{1}{c}{0.861} &
  \multicolumn{1}{c}{0.956} &
  \multicolumn{1}{c}{0.056} &
  \multicolumn{1}{c}{0.265} &
  \multicolumn{1}{c}{\textbf{0.001}} &
  \multicolumn{1}{c}{0.972} &
  \multicolumn{1}{c}{0.071} &
  \multicolumn{1}{c}{30.721} &
  0.971 \\
{MFIF GAN} &
  \multicolumn{1}{c}{0.043} &
  \multicolumn{1}{c}{0.597} &
  \multicolumn{1}{c}{0.865} &
  \multicolumn{1}{c}{0.948} &
  \multicolumn{1}{c}{0.070} &
  \multicolumn{1}{c}{{\ul 0.264}} &
  \multicolumn{1}{c}{\textbf{0.001}} &
  \multicolumn{1}{c}{0.970} &
  \multicolumn{1}{c}{0.071} &
  \multicolumn{1}{c}{31.050} &
  0.969 \\
{MSI DTrans} &
  \multicolumn{1}{c}{0.086} &
  \multicolumn{1}{c}{1.541} &
  \multicolumn{1}{c}{0.720} &
  \multicolumn{1}{c}{0.868} &
  \multicolumn{1}{c}{0.115} &
  \multicolumn{1}{c}{0.325} &
  \multicolumn{1}{c}{0.005} &
  \multicolumn{1}{c}{0.927} &
  \multicolumn{1}{c}{0.117} &
  \multicolumn{1}{c}{25.203} &
  0.923 \\
{SESFuse} &
  \multicolumn{1}{c}{{\ul 0.036}} &
  \multicolumn{1}{c}{0.594} &
  \multicolumn{1}{c}{{\ul 0.882}} &
  \multicolumn{1}{c}{{\ul 0.962}} &
  \multicolumn{1}{c}{{\ul 0.048}} &
  \multicolumn{1}{c}{0.256} &
  \multicolumn{1}{c}{\textbf{0.001}} &
  \multicolumn{1}{c}{{\ul 0.976}} &
  \multicolumn{1}{c}{{\ul 0.065}} &
  \multicolumn{1}{c}{{\ul 31.199}} &
  {\ul 0.976} \\
{SwinFusion} &
  \multicolumn{1}{c}{0.129} &
  \multicolumn{1}{c}{2.108} &
  \multicolumn{1}{c}{0.603} &
  \multicolumn{1}{c}{0.782} &
  \multicolumn{1}{c}{0.216} &
  \multicolumn{1}{c}{0.356} &
  \multicolumn{1}{c}{0.005} &
  \multicolumn{1}{c}{0.851} &
  \multicolumn{1}{c}{0.149} &
  \multicolumn{1}{c}{24.387} &
  0.862 \\
{VAEEDOF} &
  \multicolumn{1}{c}{\textbf{0.014}} &
  \multicolumn{1}{c}{\textbf{0.280}} &
  \multicolumn{1}{c}{\textbf{0.912}} &
  \multicolumn{1}{c}{\textbf{0.975}} &
  \multicolumn{1}{c}{\textbf{0.041}} &
  \multicolumn{1}{c}{\textbf{0.161}} &
  \multicolumn{1}{c}{\textbf{0.001}} &
  \multicolumn{1}{c}{\textbf{0.979}} &
  \multicolumn{1}{c}{\textbf{0.035}} &
  \multicolumn{1}{c}{\textbf{33.114}} &
  \textbf{0.986} \\
  \bottomrule
\end{tabular}%
}
\end{table*}
\Cref{tab:results} provides a summary of the quantitative evaluation, highlighting the superior performance of our method on all metrics considered. Since we have \GT, we used image quality assessment metrics to evaluate the results.

Given that different regions of the image present varying levels of difficulty for fusion models, we further analyze how performance varies when faced with more challenging areas. To do this, we divided the image into smaller patches and computed a difficulty score for each patch. This score was assigned by comparing the \GT patch with the corresponding input patches, capturing the disparity between the available information and the ideal fusion outcome. After assigning difficulty scores, we evaluated the fusion performance for each patch by computing the relevant metrics. These results show that our model exhibits greater resilience in challenging regions. Our approach maintains a more consistent performance in patches with varying levels of missing information (\Cref{fig:scores}). In contrast, other methods struggle with handling harder regions. However, it is important to note that, unlike masking-based methods, achieving maximum pixel-level similarity is not possible with our approach. This is because the encoder-decoder architecture introduces minimal discrepancies, which, although present, are largely unnoticeable to the naked eye, as shown in \Cref{fig:vae_err}.



\begin{figure*}[!h]
    \centering
    \includegraphics[width=\textwidth]{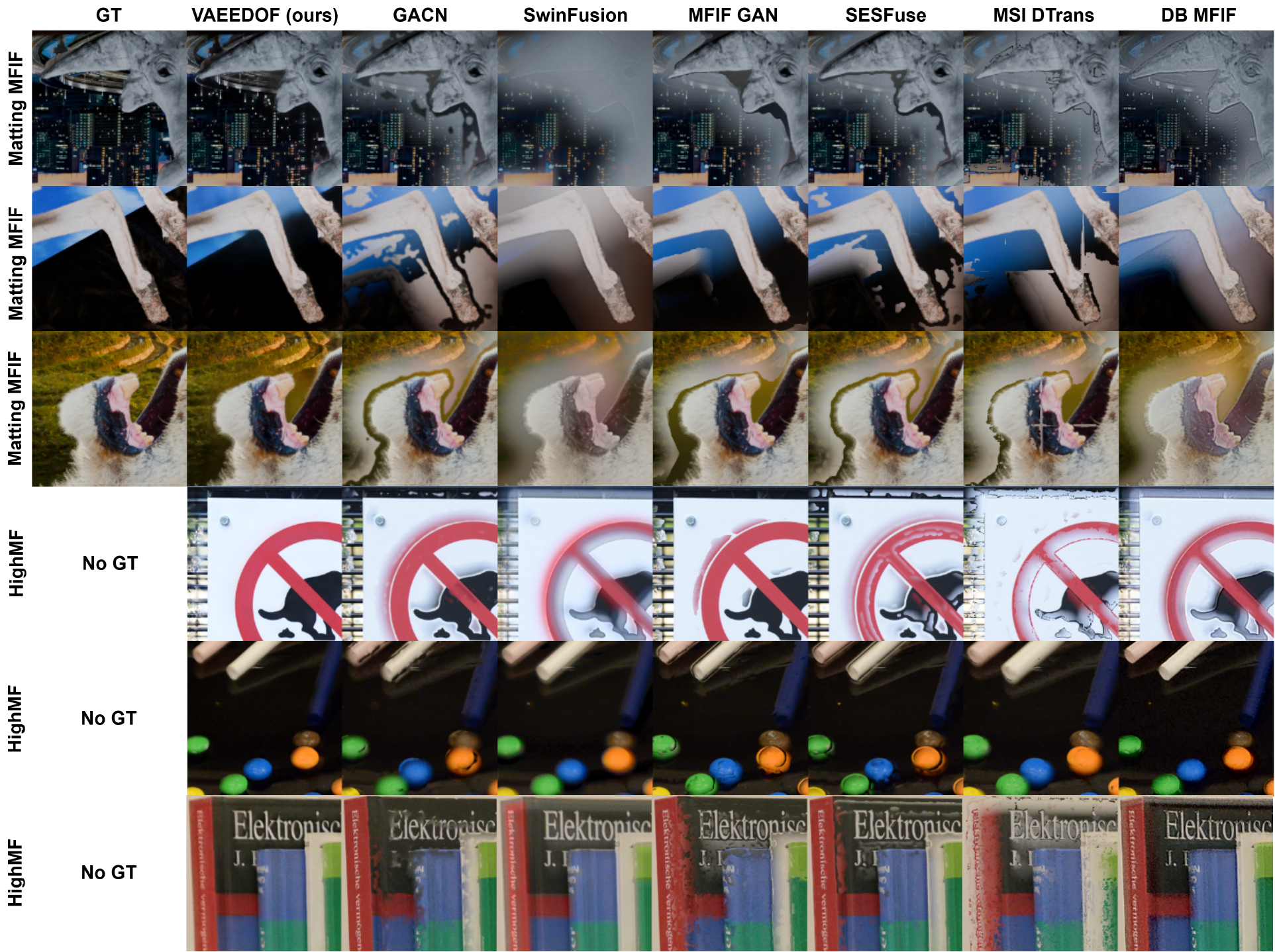}
    \caption{The figure displays the output of all models under consideration. The first three rows are from our proposed dataset, while the remaining three are from the HighMF\tcite{xiao2022dmdn} dataset. The images are selected to provide a broad representation of each model’s performance.}
    \label{fig:results}
\end{figure*}

\Cref{fig:results} compares visual results of the various methods. Our approach generates results with fewer artifacts and outperforms state-of-the-art methods in terms of quality. A key advantage of our method is that even in areas with missing information, the added content blends seamlessly, without drawing attention to itself. This results in a more natural and cohesive output compared to other methods, where artifacts in such regions are more noticeable. In addition, to evaluate the generalizability of our method, we tested them on the HighMF\tcite{xiao2022dmdn} dataset. Despite the dataset containing images that are not perfectly registered, VAEEDOF outperforms other methods, although it is not specifically trained for this type of input. 

\begin{figure}[!h]
    \centering
    \includegraphics[width=\columnwidth]{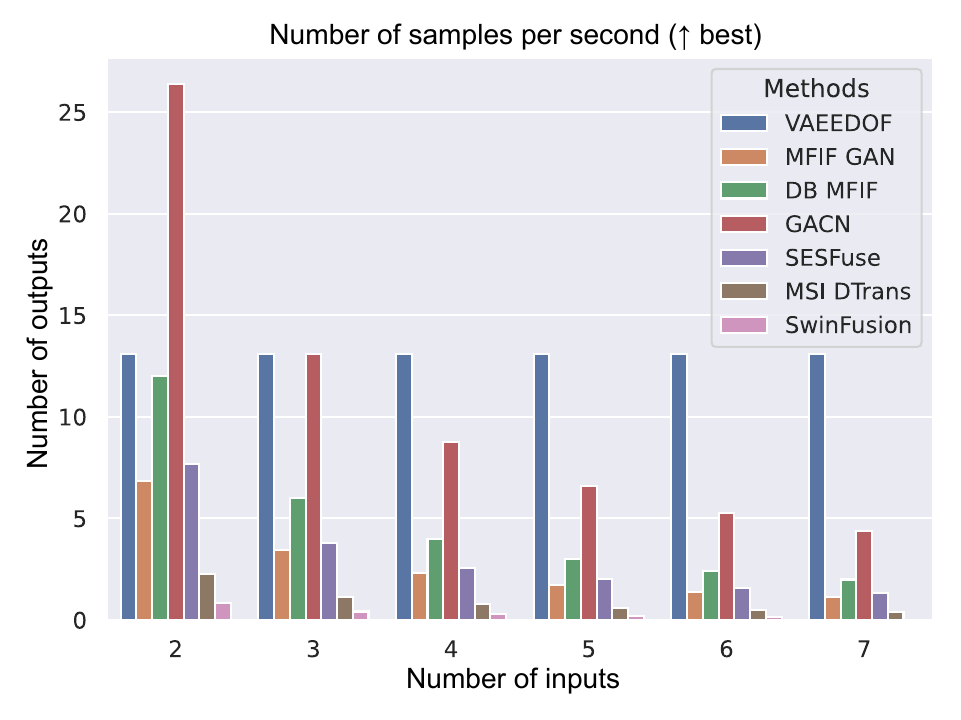}
    \caption{Bar plot showing the number of images generated (512x512) per second for various models as the number of input images increases. Higher values indicate better performance, with our method achieving significantly faster processing times as the number of input images grows, outperforming the other models in terms of efficiency.}
    \label{fig:time}
\end{figure}

Finally, our approach exhibits remarkable speed. When merging only two images, our method outperforms all others except the GACN\tcite{ma2022end} method. However, as the number of images to be fused increases, our method demonstrates a processing speed enhancement of up to three times greater than the next faster technique (\Cref{fig:time}).

\section{Limitations}
\label{sec:limitations}

Despite its strong performance, our method has some limitations. It assumes that the input images are perfectly registered, which may restrict its applicability in situations involving unaligned images. Although our experiments on the HighMIF dataset (\Cref{fig:results}) demonstrate that the method still outperforms state-of-the-art approaches even in the presence of moderate misalignment, the current model is not specifically designed to handle such cases.
Additionally, our two-stage architecture, while beneficial in enhancing results, preventing overfitting, and bridging the domain gap between synthetic and real-world images, introduces an additional point of failure. The quality of the final output is intrinsically tied to the performance of the \VAE ; if the \VAE introduces artifacts or significantly distorts the input, the fused output will inevitably suffer. Fortunately, this issue can be easily diagnosed by passing the input through the \VAE and measuring the reconstruction error, which offers a practical safeguard during deployment.
\section{Conclusion}
\label{sec:conclusion}

In this work, we proposed a novel approach for \MFIF, addressing the limitations of existing methods by leveraging a frozen pre-trained \VAE, a robust U-Net architecture, and an overlapping patch strategy. Our method is specifically designed to overcome the challenges posed by traditional masking-based techniques, which struggle in regions lacking information, and generative models, which often introduce artifacts and color shifts. Moreover, to train and evaluate our model, we introduced Matting\MFIF, a new 4K dataset generated using Blender, which closely simulates real-world \DOF effects while providing perfect ground truth, enhancing the diversity and complexity of fusion challenges, bridging the gap between synthetic and real-world scenarios.

Extensive experiments show that VAEEDOF achieves state-of-the-art performance on both our evaluation dataset and the HighMF dataset. Additionally, our approach is computationally efficient, outperforming other methods as the number of input images increases. However, our method currently assumes that input images are perfectly registered, which may limit its applicability in scenarios involving unaligned or moving inputs. Future work will focus on enhancing the robustness of our approach by introducing camera movement to the dataset, enabling the model to achieve better results on unregistered images.

In summary, our method presents a significant step forward in \MFIF, offering a robust and efficient solution for generating high-quality all-in-focus images.

Enhanced image fusion has the potential to improve clarity and interpretability in domains such as medical imaging, microscopy, surveillance, and digital photography. These advancements could benefit society significantly, but also raise privacy and ethical concerns, particularly in surveillance contexts, where increased image fidelity could facilitate unauthorized identification or tracking. Although our current work does not directly enable such applications, we acknowledge these risks and emphasize the importance of integrating ethical safeguards and robust misuse detection methods in future research.


\bibliography{references}
\bibliographystyle{plain}

\clearpage

\end{document}